\documentclass[conference]{IEEEtran}
\usepackage{caption}
\usepackage{multirow}
\usepackage{lscape}
\usepackage{booktabs, makecell} 
\usepackage{geometry}
\usepackage{pdflscape} 
\usepackage{graphicx}
\usepackage{xtab}
\usepackage{amsmath}
\usepackage{algorithm}
\usepackage{algpseudocode}
\usepackage{enumitem}
\usepackage{tikz}
\usepackage{subcaption}

\usetikzlibrary{shapes.multipart, arrows.meta, positioning}

\usepackage{cite}
\title{Fair Learning for Bias Mitigation and Quality Optimization in Paper Recommendation}
\author{
  \IEEEauthorblockN{%
    Uttamasha Anjally Oyshi, Susan Gauch\,}
  \IEEEauthorblockA{%
    Department of Electrical Engineering \& Computer Science\\
    University of Arkansas\\
    Fayetteville, USA\\
    e-mails: {\tt \{uoyshi, sgauch\}@uark.edu}
} }
\begin{document}
\maketitle
\begin{abstract}
Despite frequent double-blind review, demographic biases of authors still disadvantage the underrepresented groups. We present Fair-PaperRec, a MultiLayer Perceptron (MLP) based model that addresses demographic disparities in post-review paper acceptance decisions while maintaining high-quality requirements. Our methodology penalizes demographic disparities while preserving quality through intersectional criteria (e.g., race, country) and a customized fairness loss, in contrast to heuristic approaches. Evaluations using conference data from ACM Special Interest Group on Computer-Human Interaction (SIGCHI), Designing Interactive Systems (DIS), and Intelligent User Interfaces (IUI) indicate a 42.03\% increase in underrepresented group participation and a 3.16\% improvement in overall utility, indicating that diversity promotion does not compromise academic rigor and supports equity-focused peer review solutions.
\end{abstract}

\section{Introduction}
Double-blind review often does not eradicate systemic biases linked to authors’ demographics, reputations, or institutional affiliations, despite attempts to ensure impartiality \cite{tomkins2017reviewerdup, shmidt2022double, giannakakos2024impact, mebane2025double}. Recent data indicates that even the most stringent anonymization techniques can be undermined by analyzing writing style or cross-referencing previous articles \cite{bauersfeld2023cracking, shah2023role}. This tendency can sustain biases against particular groups, including women, racial minorities, and researchers from underrepresented areas \cite{huber2022nobel, frachtenberg2022metrics, lee2013biasdup, giannakakos2024impact}. Simultaneously, there is a growing dependence on recommendation algorithms to optimize processes such as paper selection, grant distribution, and significant publication identification \cite{goues2017effectiveness, bobadillaDeepFair, peng2023reranking}. While these systems can accelerate decision-making, they also pose a danger of perpetuating biases present in the training data, particularly if they focus only on predictive accuracy \cite{morik2020controlling, beutel2017data, yao2017beyond}. Therefore, it is imperative to devise novel methodologies that explicitly include demographic justice, preventing the perpetuation of historical inequalities.

In this paper, we introduce Fair-PaperRec, a fairness-aware recommendation framework specifically designed to mitigate post-review bias. Unlike previous heuristic-based approaches that often handle single-attribute fairness constraints or overlook intersectionality, in our approach:

\begin{figure*}[ht]
    \captionsetup{font={footnotesize,rm},justification=centering,labelsep=period}%
    \centering
    \includegraphics[width=0.95\linewidth]{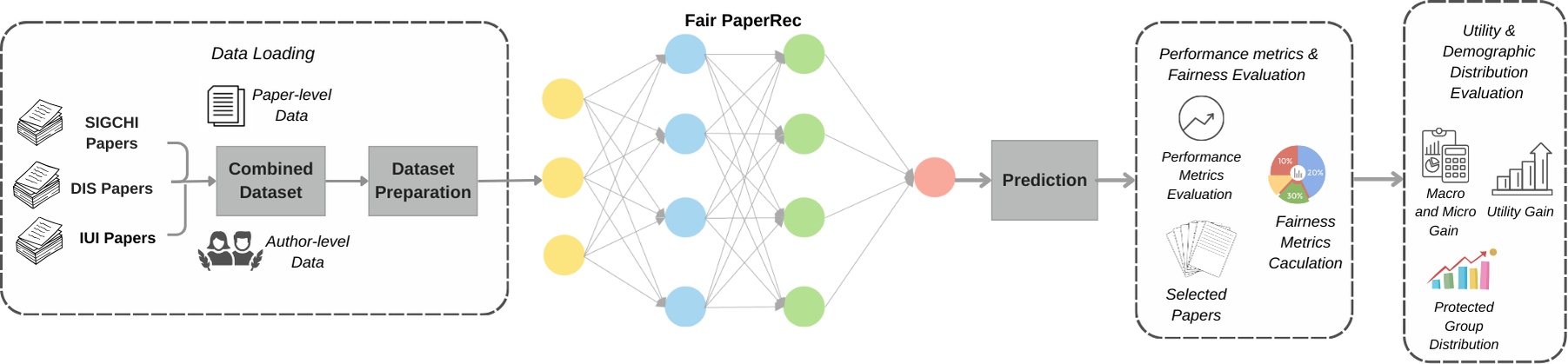}
    \caption{Overview of the Fair-PaperRec Architecture.}
    \label{fig:arch}
    \vskip -10pt
\end{figure*}
\captionsetup{font={footnotesize,rm},justification=centering,labelsep=period}%

\begin{itemize}
    \item  We surpass single-attribute approaches by incorporating multiple demographic attributes (e.g., race, country) and constructing multi-dimensional profiles that capture underlying biases.
    \item  After a double-blind review, a specialized fairness penalty is implemented to address demographic disparities, thereby correcting latent biases without the need to replace existing processes.
    \item  Our method ensures that the quality of the paper is maintained throughout by ensuring demographic parity, thereby obtaining equitable representation without compromising academic rigor.
\end{itemize}

Our results demonstrate improved representation in the participation of underrepresented groups, as well as an enhancement in overall paper quality, as indicated by the h-index. Notably, these findings reveal that enhanced inclusivity need not diminish academic rigor; a fairness-driven approach can yield greater demographic parity while simultaneously preserving, and at times even \emph{enhancing}, the quality of accepted papers.

By mitigating biases in paper selection, our strategy promotes a richer academic discourse and amplifies the representation of marginalized communities, thereby paving the way toward more equitable, high-quality conferences. The paper includes the folling sections, where in Section 2, we review related work. Section 3 presents the proposed methodology. Section 4 explains our experimental setup and metrics. Section 5 provides results and analysis. Finally, the Section 6 concludes the paper.

\section{Related Work}
We begin by examining double-blind review and bias in academic paper selection, then explore fairness in
recommender systems, and finally discuss recent advancements in neural approaches for fair selection.

\subsection{Double-Blind Review and Bias in Academic Paper Selection}
Although double-blind review conceals identities \cite{tomkins2017reviewerdup,shmidt2022double,giannakakos2024impact}, it often fails to eliminate biases in gender, race, or geography \cite{lee2013biasdup,bol2023geographical}. While authorship-attribution can rectify advanced anonymization \cite{bauersfeld2023cracking}, high-prestige institutions continue to receive favorable reviews \cite{lee2013biasdup2}. As a result, underrepresented groups, including women and racial minorities, continue to be marginalized \cite{williams_national_2015}, and substantial acceptance rate disparities persist \cite{huber2022nobel,frachtenberg2022metrics}.

\subsection{Fairness in Recommendation Systems}

When optimizing solely for accuracy, recommenders frequently exacerbate biases \cite{burke2017multisided,bobadillaDeepFair}. Although some fairness issues are addressed by multi-objective \cite{morik2020controlling}, adversarial \cite{beutel2017data}, and re-ranking methods \cite{yao2017beyond}, the majority of these methods concentrate on single attributes or user-item data, leaving intersectional biases in paper acceptance unaccounted for. In academic settings, \emph{provider fairness} is equivalent to \emph{author fairness}, which protects minority researchers \cite{alsaffar2021multidimensional}. There are very few algorithms that resolve post-review bias, not to mention, multi-attribute fairness \cite{peng2023reranking,fu2020fairnessaware}.

\subsection{Post-Review Bias Mitigation and Neural Approaches}
Some heuristic methods attempt to rebalance accepted papers after reviews \cite{alsaffar2021multidimensional}, but they risk local optima and often fail to consider multi-attribute fairness. Neural-based solutions such as \emph{DeepFair} \cite{bobadillaDeepFair} or \emph{Neural Fair Collaborative Filtering} \cite{islam2021neural} demonstrate that fairness can align with accuracy, yet they typically target commercial recommendations rather than the nuances of academic peer review. Meanwhile, multi-stakeholder optimization \cite{wu2022multi, wang2023survey} highlights the need for more contextual fairness definitions within scholarly publishing. Although certain approaches (e.g., Bulut et al. \cite{bulut2018paper}) employ text-based features like Term Frequency–Inverse Document Frequency (TF-IDF) to improve relevance, they often disregard the imperative of equity for authors from historically marginalized groups.

\section{Methodology}
Our approach tackles demographic biases in conference data by employing a simple Multilayer Perceptron (MLP) to enforce fairness post-review. We highlight two fundamental principles: (1) revealing and alleviating biases instead of eliminating them, and (2) implementing a straightforward, yet efficient neural architecture that harmonizes equality and utility.

\subsection{Data Collection and Pre-processing}
Real-world datasets—particularly those drawn from academic conference submissions—often contain latent biases that mirror systemic imbalances in the scholarly community (e.g., underrepresentation of certain demographics). We utilize datasets from SIGCHI 2017, DIS 2017, and IUI 2017~\cite{alsaffar2021multidimensional}, which naturally reflect systemic disparities (e.g., skewed demographics). Instead of eliminating such biases, our objective is to \emph{recognize and rectify} them.

We describe the process of collecting and preparing the data used in our experiments. The dataset consists of academic papers submitted to conferences, and we employ a variety of pre-processing steps to ensure the data are suitable for training our model.

\begin{table}[ht]
\centering
\captionsetup{font={footnotesize,rm},justification=centering,labelsep=period}%
\caption{\MakeUppercase{Demographic Participation from protected groups in Three Conferences.}}
\label{table:demographics}
\begin{tabular}{lccc}
    \toprule
    \textbf{Conference} & \textbf{Gender (\%)} & \textbf{Race (\%)} & \textbf{Country (\%)} \\
    \midrule
    SIGCHI & 41.88 &   6.84 &    21.94 \\
    DIS    & 65.79 &  35.09 &    24.56 \\
    IUI    & 43.75 &  51.56 &    39.06 \\
    \midrule
    Average & 50.47 &  31.16 &    28.52 \\
    \bottomrule
\end{tabular}
\end{table}
\captionsetup{font={footnotesize,rm},justification=centering,labelsep=period}%

\subsubsection{Data Description}
We gathered detailed information at the paper and author levels, resulting in a robust combined dataset. Every paper record has a title, authors, and a conference designation (1 = IUI, 2 = DIS, 3 = SIGCHI). Author records encompass demographic information (gender, race, nationality, career stage), for detailed analysis. We classify SIGCHI 2017 articles as a standard for high-impact research, whereby \emph{Overall} includes all submissions and \emph{Selected} refers to those identified by our algorithms.

\subsubsection{Data Pre-processing}
Several preprocessing steps were undertaken to prepare the dataset for training: 

\begin{itemize}
    \item \emph{Categorical Encoding:} Gender, Country, and Race are subjected to one-hot encoding. Gender is binary (0 = male, 1 = female), Country is categorized as \emph{developed} or \emph{underdeveloped}, and Race comprises \{White, Asian, Hispanic, Black\}, with Hispanic and Black designated as protected groups (Table~\ref{table:demographics}).
    \item \emph{Normalization:} Numerical attributes (e.g., h-index) employ min-max scaling for consistent magnitude.
    \item \emph{Training and Validation Division:} An 80\%/20\% stratified division guarantees equitable distribution of labels and protected attributes in both subsets.
\end{itemize}

\subsection{Problem Definition}

This study develops a \emph{fairness-aware paper recommendation system} that ensures demographic parity with respect to authors’ race and country, while preserving high academic standards. We frame acceptance decisions as a \emph{recommendation} task, where \emph{conference organizers (users)} seek to select from \emph{530 papers (items)} spanning SIGCHI, DIS, and IUI. Each paper (item) includes an \emph{h-index} for quality, demographic data (race, country), and a conference rating (SIGCHI: 1, DIS: 2, IUI: 3).

Our approach enforces fairness constraints on race and country independently, excluding \emph{gender} due to its relatively balanced distribution (see Table~\ref{table:demographics}). By leveraging historical acceptance patterns and explicit diversity goals, the system balances the \emph{need for high-quality research} with the \emph{requirement to address demographic biases} in the final recommendation of papers.

Let \( D \) represent the dataset of submitted papers, where each paper \( p \in D \) is associated with a set of features \( X_p \) (e.g., race, country, h-index) and a target variable \( y_p \) indicating acceptance (1) or rejection (0). The \emph{race} attribute \( R_p \) and \emph{country} attribute \( C_p \) are the protected attributes.

We aim to optimize a predictive model \( f: X_p \rightarrow \hat{y}_p \) that minimizes the following objective function:

\begin{equation}
    \min_{f} \left( \mathcal{L}(f(X_p), y_p) + \lambda \cdot \mathcal{L}_{\text{fairness}}(f, D) \right)
    \label{eq:eq_1}
\end{equation}

Here, \( \mathcal{L}(f(X_p), y_p) \) is the \emph{prediction loss} (e.g., Binary Cross-Entropy Loss), \( \mathcal{L}_{\text{fairness}}(f, D) \) is the \emph{fairness loss}, penalizing deviations from demographic parity across race and country and \( \lambda \) is a hyperparameter that balances the trade-off between prediction accuracy and fairness.

\begin{algorithm}
\captionsetup{font={footnotesize,rm},justification=centering,labelsep=period}%
\caption{\MakeUppercase{Fair-PaperRec Loss Function.}}
\footnotesize
\label{alg:algorithm1}
\begin{algorithmic}[1]
    \State \textbf{Input}: Model \( M \), Epochs \( E \), Batch size \( B \), Data \( D \), Protected attributes \( A \), Hyperparameter \( \lambda \)
    \State \textbf{Output}: Trained Model \( M \)
    \State \textbf{Initialize} Model \( M \)
    \For{each \( e \in E \)}
        \State Shuffle Data \( D \)
        \For{each batch \( \{ (X, Y) \} \in D \) with size \( B \)}
            \State Predict \( \hat{Y} \leftarrow M(X) \)
            \State Calculate Loss:
            \State \quad \( L_{\text{prediction}} \leftarrow \text{PredictionLoss}(Y, \hat{Y}) \)
            \State \quad \( L_{\text{fairness}} \leftarrow \text{FairnessLoss}(A, \hat{Y}) \)
            \State Calculate Total Loss:
            \State \quad \( L_{\text{total}} \leftarrow \lambda \cdot L_{\text{fairness}} + L_{\text{prediction}} \)
            \State Compute gradients \( \nabla L_{\text{total}} \leftarrow \frac{\partial L_{\text{total}}}{\partial M} \)
            \State Update Model parameters: \( M \leftarrow M - \alpha \nabla L_{\text{total}} \)
        \EndFor
    \EndFor
\end{algorithmic}
\end{algorithm}
\captionsetup{font={footnotesize,rm},justification=centering,labelsep=period}%

\subsection{Demographic Parity}
We aim to ensure that the probability of a paper being accepted is independent of the protected attributes:
\begin{equation*}
    P(\hat{y}_p = 1 \mid R_p = r) = P(\hat{y}_p = 1), \quad \forall r \in \text{Race}
    \label{eq:eq_2}
\end{equation*}
\begin{equation*}
    P(\hat{y}_p = 1 \mid C_p = c) = P(\hat{y}_p = 1), \quad \forall c \in \text{Country}
    \label{eq:eq_3}
\end{equation*}

Utilizing these equations ensures that the papers authored by individuals from different races and countries have an equal probability of acceptance.

\subsection{Fairness Loss}
The fairness loss from the objective function in Equation \ref{eq:eq_1} is constructed to minimize statistical parity differences between the protected and non-protected group:
\begin{equation}
    \mathcal{L}_{\text{fairness}} = \left( P(\hat{y}_p = 1 \mid G_{\text{p}}) - P(\hat{y}_p = 1 \mid G_{\text{np}}) \right)^2
    \label{eq:eq_4}
\end{equation}

Here, \( P(\hat{y}_p = 1 \mid G_{\text{p}}) \) denotes the acceptance probability for the protected group and \( P(\hat{y}_p = 1 \mid G_{\text{np}}) \) is the acceptance probability for the non-protected group.

\subsection{Combined Fairness Loss}
Furthermore, we define a combined fairness loss to minimize statistical parity differences across race and country attributes between the protected and unprotected groups, as shown in Equation \ref{eq:combinedloss}.

\begin{equation}
\begin{split}
    \mathcal{L}_{\text{fairness}} = W_{r} 
    \left( \frac{1}{N_{r}} \sum_{p \in G_{r}} \hat{y}_p 
    - \frac{1}{N} \sum_{p=1}^{N} \hat{y}_p \right)^2 \\
    + W_{c} 
    \left( \frac{1}{N_{c}} \sum_{p \in G_{c}} \hat{y}_p 
    - \frac{1}{N} \sum_{p=1}^{N} \hat{y}_p \right)^2
\end{split}
\label{eq:combinedloss}
\end{equation}
\( G_{\text{r}} \) and \( G_{\text{c}} \) denote the race and country groups, respectively.
\( N_{r} \) and \( N_{c} \) are the number of papers in each group and weights \( W_{r} \) and \( W_{c} \) reflect group distributions.

\subsection{Total Loss}
The total loss is the combination of prediction and fairness losses:

\begin{equation*}
    \mathcal{L}_{\text{total}} = \mathcal{L}_{\text{prediction}} + \lambda \cdot \mathcal{L}_{\text{fairness}}
\end{equation*}

\begin{table}[h!]
\centering
\captionsetup{font={footnotesize,rm},justification=centering,labelsep=period}%
\caption{\MakeUppercase{Gain Calculations for Country and Race Features with Utility Gain ($UG_{\text{i}}$).}}
\renewcommand{\arraystretch}{1} 
\setlength{\tabcolsep}{2pt} 
\scriptsize 
\label{table:gaincalc}
\begin{tabular}{l c c c c c c}
    \toprule
    \multicolumn{1}{l}{} & \multicolumn{3}{c}{\textbf{Country Feature}} & 
    \multicolumn{3}{c}{\textbf{Race Feature}} \\
    \cmidrule(lr){2-4} \cmidrule(lr){5-7}
    \textbf{$\lambda$} & \makecell{\textbf{Macro} \\ \textbf{Gain (\%)}} & \makecell{\textbf{Micro} \\ \textbf{Gain (\%)}} & \makecell{\textbf{$UG_{\text{i}}$} \\ \textbf{(\%)}} & \makecell{\textbf{Macro} \\ \textbf{Gain (\%)}} & \makecell{\textbf{Micro} \\ \textbf{Gain (\%)}} & \makecell{\textbf{$UG_{\text{i}}$} \\ \textbf{(\%)}} \\
    \midrule
    1   & 7.71  & 8.67  & 3.16  & 24.81  & 31.11  & 0.35 \\
    2   & 10.77 & 13.23 & 1.05  & 33.54  & 46.30  & 1.75 \\
    2.5 & \textbf{12.67} & \textbf{22.96} & \textbf{1.75}  & 39.25  & 54.81  & 1.40 \\
    3   & 13.60 & 16.96 & 0.35  & \textbf{42.03}  & \textbf{56.48}  & \textbf{3.16} \\
    5   & 14.80 & 19.97 & -0.35 & 43.04  & 56.11  & -0.70 \\
    10  & 13.86 & 18.73 & 2.46  & 52.91  & 64.81  & -0.70 \\
    \bottomrule
\end{tabular}
\vskip -10pt
\end{table}
\captionsetup{font={footnotesize,rm},justification=centering,labelsep=period}%

\subsection{Constraints and Considerations}
We assess fairness by training our model separately on \emph{race} and \emph{country}, as well as jointly on both attributes to evaluate selection fairness across multiple dimensions.

\paragraph{Exclusion of Protected Attributes}
Race \(R_p\) and country \(C_p\) are excluded from the input feature set \(X_p\) to mitigate direct bias amplification. To achieve joint fairness, both attributes are omitted during training, preventing the model from learning acceptance outcomes influenced by race or country.

\paragraph{Indirect Bias Mitigation}
A fairness loss promotes demographic parity, addressing indirect biases associated with features related to race or country. The model maintains neutrality by penalizing selection disparities, even in the absence of protected attributes.

\paragraph{Scalability}
Our method supports datasets of varying scales and complexities, demonstrating strong performance across various academic fields. This scalability ensures fairness across various use cases.

\section{Model Overview}

To achieve demographic parity while preserving quality in paper selection, we present a MLP-based neural network (See Figure \ref{fig:arch}), explicitly engineered to balance the trade-off between fairness and accuracy. It illustrates the correlations between input features, like author demographic attributes and paper quality, while alleviating biases during selection.

A unique fairness loss function was employed to ensure equity, imposing penalties on the model for substantial differences in selection rates between protected and non-protected groups. This loss function is integrated with the conventional prediction loss to attain a balance between diversity and accuracy; the algorithm is shown in Algorithm \ref{alg:algorithm1}.

The acceptance probabilities for submitted papers are generated by the MLP, which are subsequently ranked to guarantee that the final selection meets both quality and fairness objectives. By selecting top papers according to these probabilities, we ensure equal representation of authors from both protected and non-protected groups while upholding the requisite standard of academic excellence.

\begin{algorithm}
\captionsetup{font={footnotesize,rm},justification=centering,labelsep=period}%
\caption{\MakeUppercase{Fairness-Aware Paper Selection Mechanism.}}
\footnotesize
\label{alg:algo2}
\begin{algorithmic}[1]
    \State \textbf{Input}: Dataset \( D \), Model \( M \), Number of Accepted Papers \( N_a \), Total Papers \( N_t \)
    \State \textbf{Output}: Selected Papers \( P_{\text{selected}} \)
    \State \textbf{Initialize}: \( P_{\text{selected}} \leftarrow \emptyset \)
    \State \textbf{Step 1}: Apply trained model \( M \) to the entire dataset \( D \)
    \For{each paper \( p \in D \)}
        \State Compute acceptance probability: \( \hat{y}_p \leftarrow M(p) \)
    \EndFor
    \State \textbf{Step 2}: Rank all papers \( p \) by acceptance probability \( \hat{y}_p \)
    \State Sort \( D \) in descending order of \( \hat{y}_p \)
    \State \textbf{Step 3}: Select top \( N_a \) papers:
    \State \quad \( P_{\text{selected}} \leftarrow \{ p \mid \hat{y}_p \geq \hat{y}_{(N_a)} \} \)
    \State \textbf{Step 4}: Ensure Fairness Constraints
    \State \textbf{Return} \( P_{\text{selected}} \)
\end{algorithmic}
\end{algorithm}
\captionsetup{font={footnotesize,rm},justification=centering,labelsep=period}%
\vskip -10pt

\subsection{Selection Mechanism}

The model calculates acceptance probabilities for all submitted papers after training. After calculating acceptance odds, the algorithm ranks candidate papers. This rating phase ensures underrepresented groups are represented in final admission decisions. Representing this as a suggestion list preserves the peer-review process and corrects residual biases. Algorithm \ref{alg:algo2} selects the best papers based on probability, ensuring fairness and preserving the desired number of accepted papers.
\begin{itemize}
    \item \emph{Prediction Aggregation:} The trained MLP model is applied to the entire dataset to obtain predicted acceptance probabilities \( \hat{y}_p \) for each paper. 
    \item \emph{Ranking:} Papers are ranked in descending order based on their predicted probabilities.
    \item \emph{Selection:} The papers with the highest predicted probabilities are selected for acceptance, ensuring the total number of selected papers matches the required acceptance quota.
\end{itemize}

Mathematically, the selection process is represented as:
\begin{equation*}
\text{Selected Papers} = \left\{ p \in D \mid \hat{y}_p \geq \hat{y}_{(N_a)} \right\}
\end{equation*}

Here, \( \hat{y}_{(N_a)} \) is the \( N_a \)-th highest predicted probability in the set \( \{ \hat{y}_p \mid p \in D \} \) while \( N_a \) is the total number of accepted papers and \( N_t \) is the total number of submitted papers, where \( N_a \leq N_t \).

This approach ensures that the selection process is both informed by the model's predictions and constrained to uphold demographic parity, fostering an equitable and meritocratic paper selection environment.

\section{Evaluation and Experiments}
This section presents the experimental evaluation of our proposed Fair-PaperRec model on the chosen datasets. To guide the exploration of fairness and quality in our proposed paper recommendation system, we pose the following research questions:

\begin{itemize}[leftmargin=*]
\item\emph{RQ1:} How do fairness constraints affect the overall quality (utility) of recommended papers, as measured by metrics, such as the h-index?
\item\emph{RQ2:} Does handling race and country as separate protected attributes differ from treating them jointly in terms of fairness outcomes and selection decisions?
\item\emph{RQ3:} How do varying weight assignments to multiple protected attributes (race and country) influence the trade-off between fairness and utility?
\end{itemize}

\begin{figure}[ht]
    \captionsetup{font={footnotesize,rm},justification=centering,labelsep=period}%
    \centering
    \includegraphics[width=0.9\linewidth]{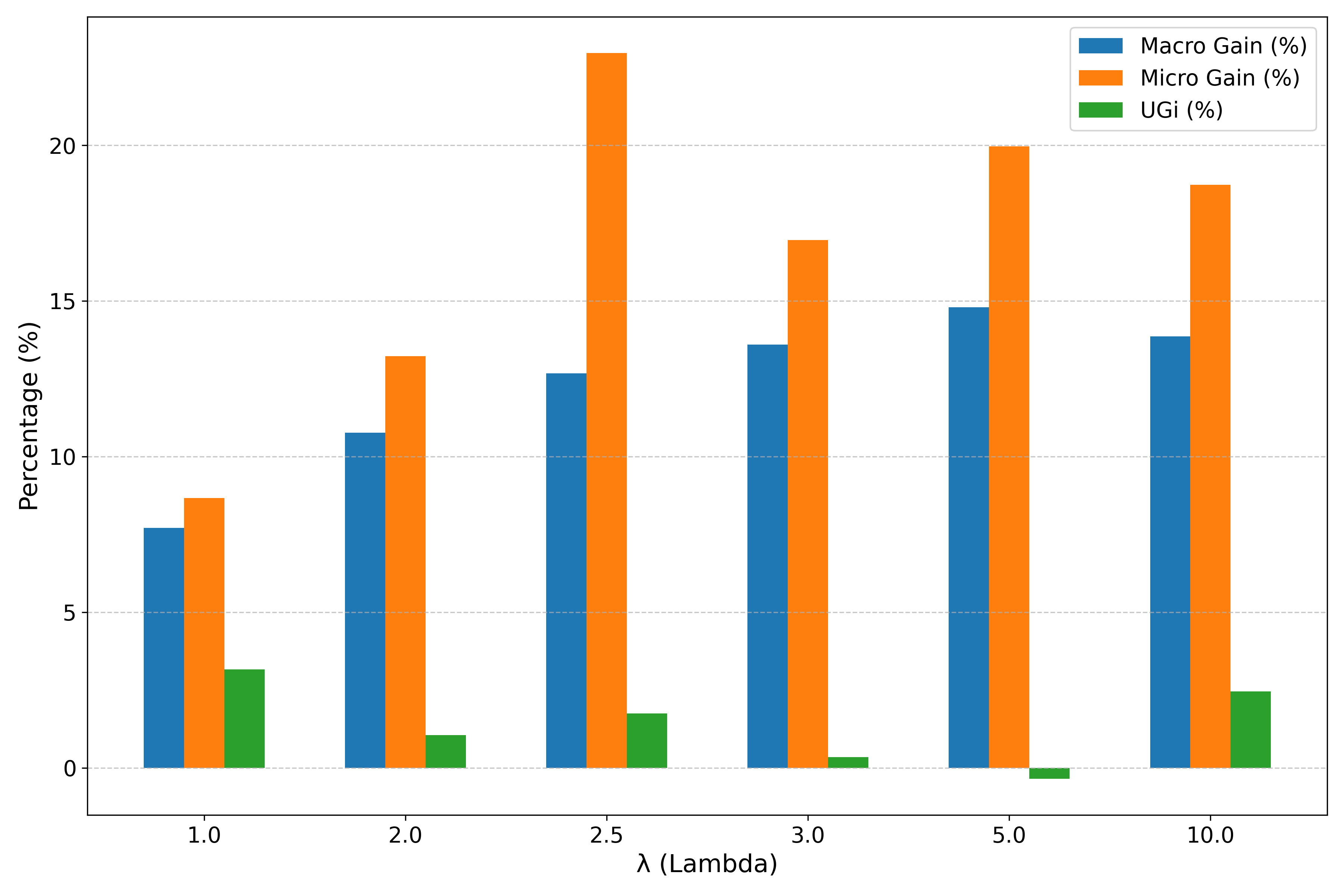} 
    \caption{Comparison of Macro and Micro Gains for Country Across Different Fairness Configurations.}
    \label{fig:countryplot}
    \vskip -10pt
\end{figure}
\captionsetup{font={footnotesize,rm},justification=centering,labelsep=period}%

\begin{figure}[ht]
    \captionsetup{font={footnotesize,rm},justification=centering,labelsep=period}%

    \centering
    \includegraphics[width=0.9\linewidth]{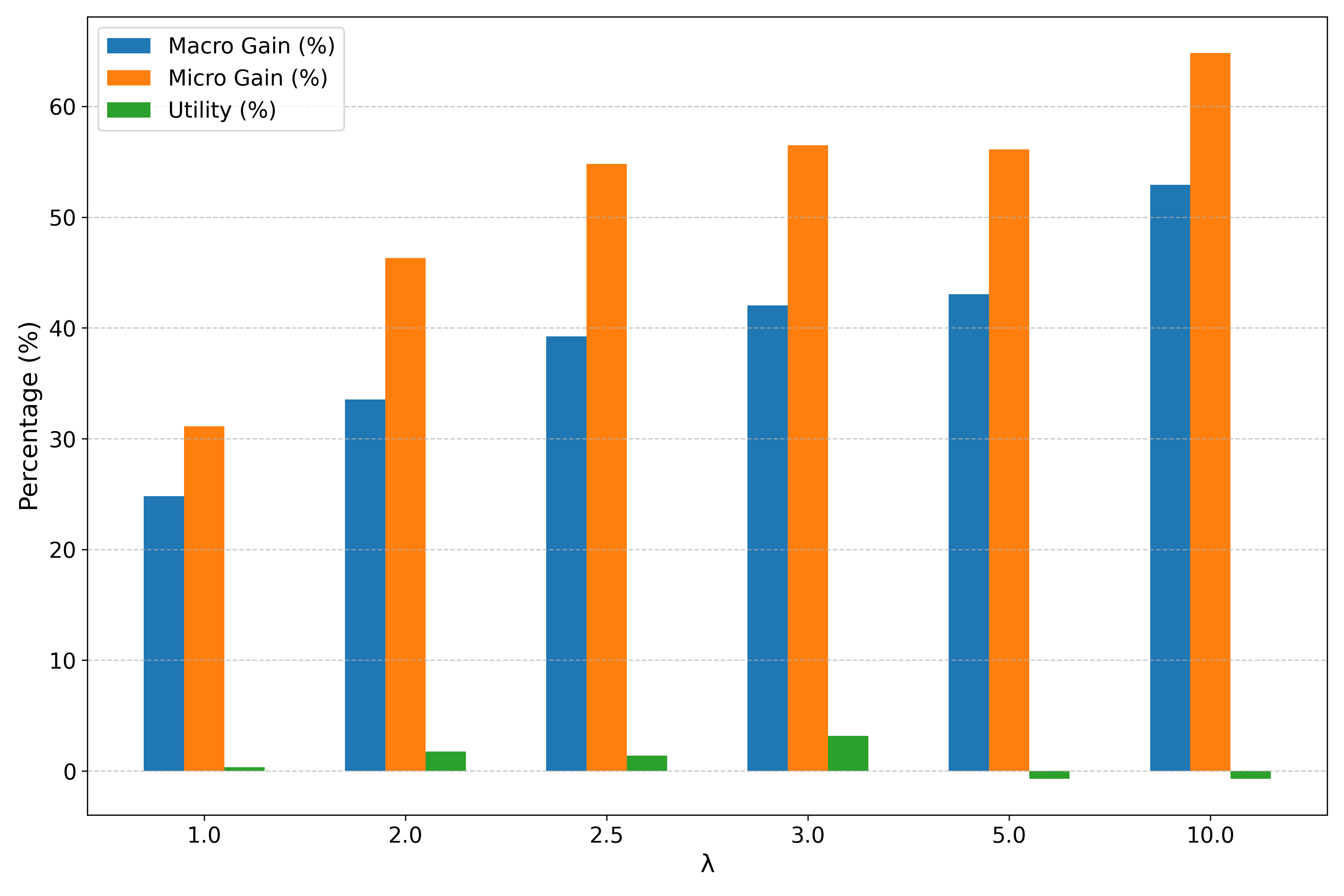} 
    \caption{Comparison of Macro and Micro Gains for Race Across Different Fairness Configurations.}
    \label{fig:raceplot}
    \vskip -10pt
\end{figure}
\captionsetup{font={footnotesize,rm},justification=centering,labelsep=period}%

\subsection{Experimental Setting}
We evaluate Fair-PaperRec using datasets from prominent academic conferences, contrasting it with baseline approaches and examining the trade-off between fairness and selection quality. Each experiment is conducted 5 times individually, with \emph{standard deviations} provided for \emph{consistency}.
\begin{table}[h!]
\centering
\captionsetup{font={footnotesize,rm},justification=centering,labelsep=period}%
\caption{\MakeUppercase{Distribution of Recommended Papers from Each Conference.}}
\begin{tabular}{lccc}
\toprule
\textbf{Label} & \textbf{Country} & \textbf{Race} & \textbf{Multi-Fair} \\
\midrule
SIGCHI  & 92.02\% & 92.00\% & 92.02\% \\
DIS     & 4.84\%  & 7.69\%  & 7.40\% \\
IUI     & 3.14\%  & 0.31\%  & 0.56\% \\
\midrule
\textbf{\# Papers} & 351 & 351 & 351 \\
\bottomrule
\end{tabular}
\vskip -10pt
\end{table}
\captionsetup{font={footnotesize,rm},justification=centering,labelsep=period}%

\subsubsection{Implementation Details}
All experiments use \texttt{PyTorch} on a high-performance machine with two  NVIDIA Quadro RTX 4000 Graphics Processing Units (GPUs). Our model is a two-hidden-layer MLP (Rectified Linear Unit (ReLU) activations, Batch Normalization), ending in a sigmoid output for acceptance probabilities. We train for 50 epochs using Adam (learning rate = 0.001), applying early stopping if no improvement occurs over 10 epochs. The fairness regularization parameter \(\lambda\) is tuned to balance utility and demographic parity. Each dataset is split 80/20 (training/validation) via stratified sampling, and each run is repeated five times with different random seeds to average performance metrics and capture variance. 

\begin{figure*}[ht]
    \captionsetup{font={footnotesize,rm},justification=centering,labelsep=period}
    \centering

    \begin{minipage}{0.32\textwidth}
        \centering
        \includegraphics[width=\linewidth]{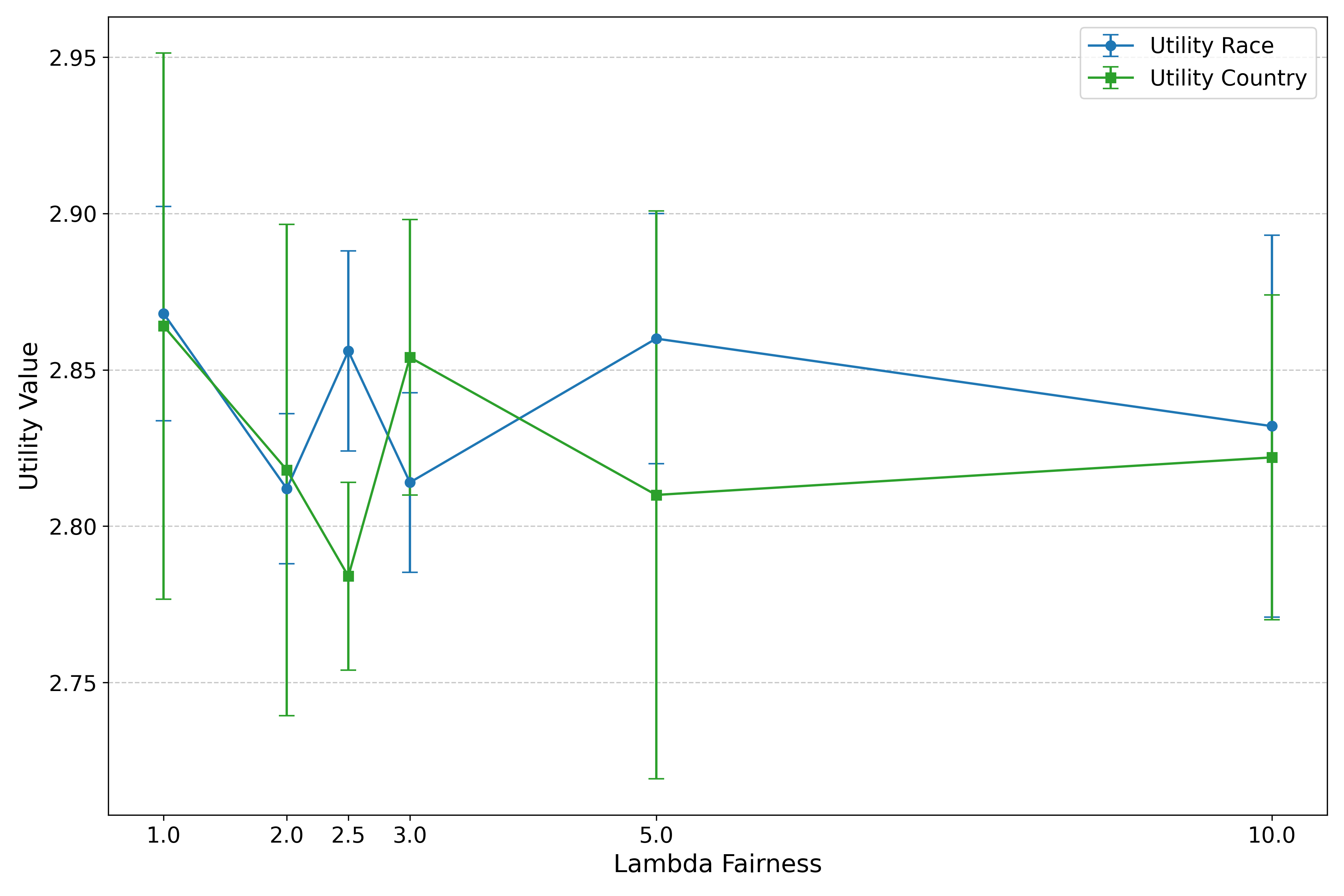}
        \caption*{(a) Utility Gain.}
    \end{minipage}
    \hfill
    \begin{minipage}{0.32\textwidth}
        \centering
        \includegraphics[width=\linewidth]{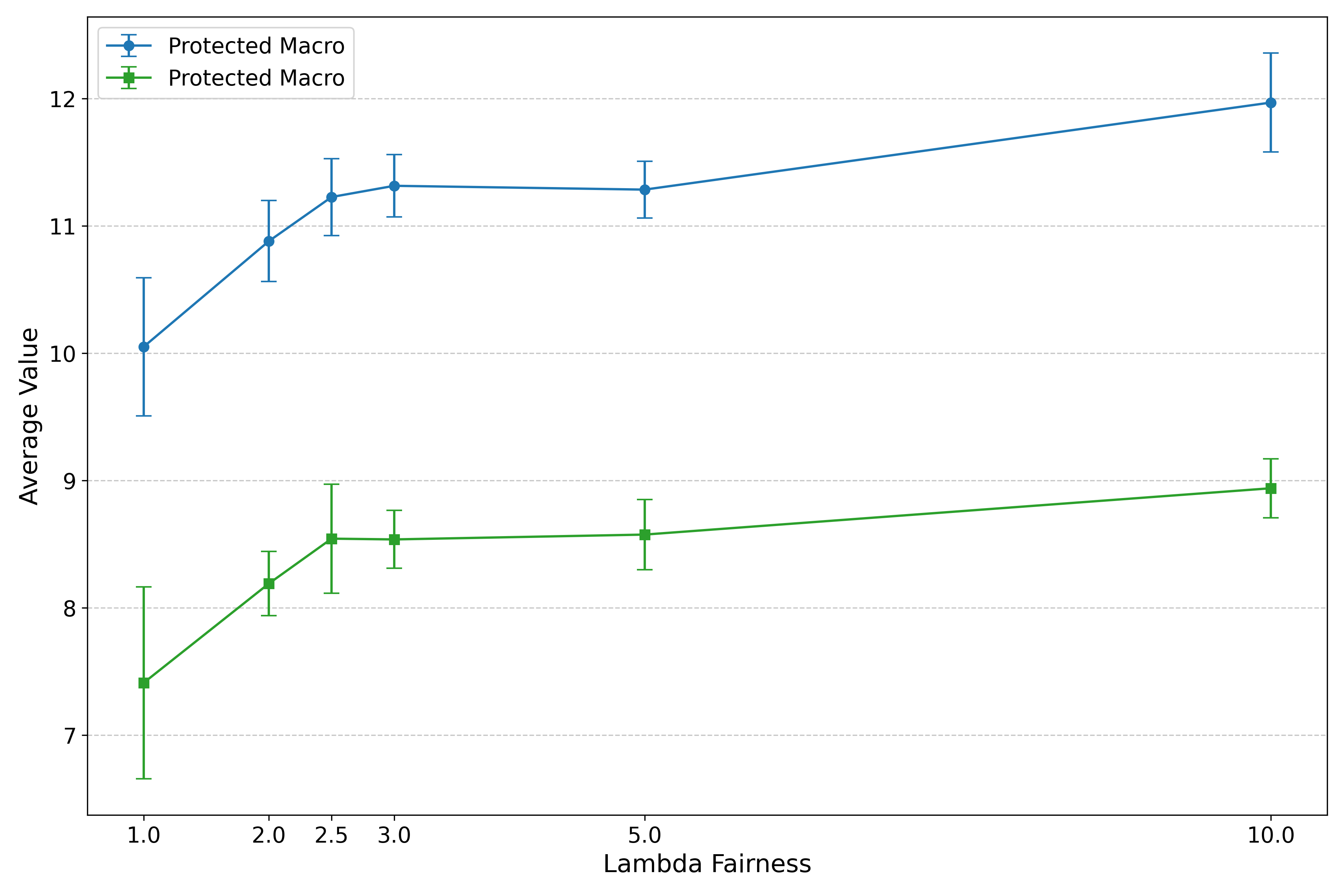}
        \caption*{(b) Macro/Micro for Race.}
    \end{minipage}
    \hfill
    \begin{minipage}{0.32\textwidth}
        \centering
        \includegraphics[width=\linewidth]{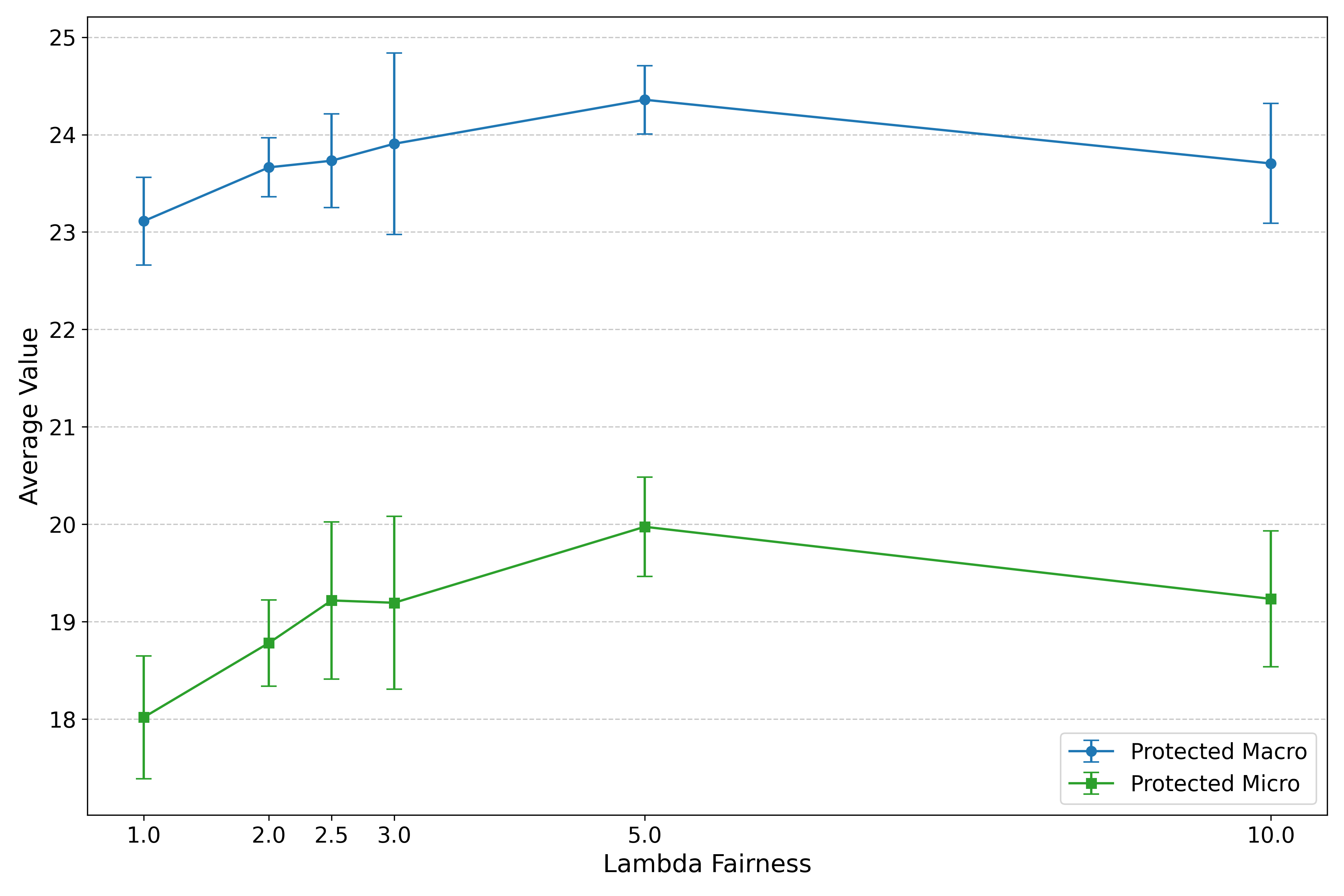}
        \caption*{(c) Macro/Micro for Country.}
    \end{minipage}

    \caption{Comparison of gains across different fairness configurations.}
    \label{fig:combinedsd}
    \vskip -10pt
\end{figure*}
\captionsetup{font={footnotesize,rm},justification=centering,labelsep=period}%

\subsubsection{Baseline}
We compare our model against a baseline Demographic-Blind Model which is a conventional (MLP) model that prioritizes quality and ignores fairness constraints. This model selects the original list of papers chosen by the SIGCHI 2017 program committee.

\subsubsection{Parameters}
A hyperparameter \(\lambda\) is used for controlling the trade-off between prediction accuracy and fairness. Higher values emphasize fairness more strongly.
    
The weights \(W_{\text{c}}\), \(W_{\text{r}}\) respectively denote the weighting factors assigned to the country and race attributes in the fairness loss function, as shown in Equation \ref{eq:combinedloss}.

\subsection{Evaluation Metrics}
Diversity is assessed at both the \emph{paper level} and the \emph{author level}. In particular:
\begin{itemize}
\item \emph{Macro Gain} represents the percentage increase in the diversity of each feature within the selected papers compared with the baseline, assessing the overall representation of protected groups.
    \item \emph{Micro Gain} is the percentage increase in the diversity of each feature among authors of the selected papers, providing more detailed perspective on inclusivity.
\end{itemize}

A \emph{Diversity Gain} \cite{alsaffar2021multidimensional} further normalizes these macro-level changes (Equation~\ref{eq:diversitygain}), capping each feature at 100 to avoid any single attribute skewing the total. The \emph{F - measure} \cite{alsaffar2021multidimensional} (Equation~\ref{eq:fmeasure}) then combines this diversity improvement with the resulting utility, offering a harmonic balance between fairness gains and paper quality.

To ensure that enhancements in diversity do not compromise the quality of papers, we assess \emph{Utility Gain} ($UG_{i}$). The utility is represented by the weighted h-index corresponding to an author's career stage—Professor, Associate Professor, Lecturer, Post-Doctoral Researcher, or Graduate Student—indicating their distribution within the dataset. Analyzing the values of the h-index in relation to a baseline determines whether equity initiatives compromise academic quality.

\begin{equation}
    D_{G} = \frac{\sum_{i=1}^{n} \min(100, \text{Macro Gain}_{G_{i}})}{n}
    \label{eq:diversitygain}
\end{equation}

\begin{equation}
    F = 2 \times \frac{D_{G} \times (100-UG_{i})}{D_{G} + (100-UG_{i})}
    \label{eq:fmeasure}
\end{equation}

\subsection{Interpretation of the Results}
The fairness regularization parameter (\(\lambda\)) was evaluated using values from 1 to 10 to examine its impact on fairness, utility, and diversity (see Table~\ref{table:gaincalc}). \emph{RQ1}, which investigates how fairness constraints affect the utility of paper recommendations, was addressed through Figures \ref{fig:countryplot} and \ref{fig:raceplot}. For the protected attribute "race," a \(\lambda\) value of 3 achieved an effective balance between diversity (both micro and macro) and utility. For "country," the optimal \(\lambda\) value was 2.5, which performed best across metrics. As \(\lambda\) increased, both micro and macro diversity gain improved, but utility decreased, indicating a reduction in the quality of recommended papers. This observation highlights the trade-off between increasing fairness and maintaining high utility, providing a clear answer to \emph{RQ1}.

The varying optimal \(\lambda\) values for race and country reflect the different disparity ratios between these protected groups. This directly addresses \emph{RQ2}, which examines how independent consideration of race and country affects fairness outcomes. The higher disparity ratio for race, which results from the smaller fraction of protected racial groups in the initial pool, requires a higher \(\lambda\) to achieve a balance between fairness and utility compared to country. Adjusting \(\lambda\) based on the specific levels of disparity in each protected group is essential to achieving optimal results. Overall, fairness interventions led to positive diversity outcomes in both micro and macro measures compared to the baseline, indicating the benefit of targeted fairness constraints.

Figure~\ref{fig:combinedsd} presents three comparisons: (a) Utility Gain, (b) Race Fairness, and (c) Country Fairness, providing insights into utility values and diversity indicators across various \(\lambda\) values. The first graph shows that utility remained relatively stable for race but fluctuated significantly for country, especially at higher \(\lambda\) values, with larger error bars indicating greater uncertainty. Utility tended to decrease for both attributes as \(\lambda\) increased, further emphasizing the trade-off between fairness and utility discussed in \emph{RQ1}.

The second and third graphs, which illustrate the protected macro and micro diversity measures for race and country, reveal that increasing \(\lambda\) consistently improved macro diversity for both attributes, with race showing more steady growth. In contrast, micro diversity measures, particularly for country, displayed more variability and less predictable improvement. These results suggest that macro diversity benefits are easier to achieve under higher fairness constraints, while micro-level improvements, especially for country, may require more targeted interventions. This finding is relevant to \emph{RQ2}, as it highlights the differential effects of fairness interventions across protected attributes and the need for careful calibration of fairness constraints.

In summary, the results indicate a clear trade-off between fairness (as measured by micro and macro diversity gains) and utility, with the optimal \(\lambda\) values differing between race and country. This suggests that fairness policies should be tailored to the specific characteristics of each protected group to balance equity and quality effectively.

Table~\ref{table:gaincalc} presents the percentage of recommended papers from SIGCHI, DIS, and IUI across various fairness constraints. Regardless of the application of country-only, race-only, or multi-attribute fairness, SIGCHI papers maintain a dominant acceptance rate of approximately 92\%, indicative of their elevated baseline acceptance rates. DIS and IUI contribute a modest but significant share of recommendations, suggesting that while SIGCHI retains prominence, the fairness constraints facilitate the inclusion of papers from smaller conferences without substantially affecting the overall distribution.

\begin{table*}[h!]
    \centering
    \captionsetup{font={footnotesize,rm},justification=centering,labelsep=period}%
    \caption{\MakeUppercase{Gain Calculations for Country and Race Features with Utility Gain.}}
    \label{table:ablationstudy}
    \renewcommand{\arraystretch}{1} 
    \setlength{\tabcolsep}{10pt} 
    \resizebox{\textwidth}{!}{ 
    \begin{tabular}{c c c c c c c c c}
        \toprule
        \multirow{2}{*}{$\lambda$} & \multirow{2}{*}{Weights} & \multicolumn{2}{c}{\textbf{Country Feature}} & \multicolumn{2}{c}{\textbf{Race Feature}} & \multirow{2}{*}{$UG_i$ (\%)} & \multirow{2}{*}{Avg.\ $D_G$ (\%)} & \multirow{2}{*}{Avg. F (\%)} \\
        \cmidrule(lr){3-4} \cmidrule(lr){5-6}
        & & Macro Gain (\%) & Micro Gain (\%) & Macro Gain (\%) & Micro Gain (\%) & & & \\
        \midrule
          & \( W_{\text{r}} = 0.32, W_{\text{c}} = 0.68 \)   & 6.17  & 6.34  & 30.51  & 46.30  & 3.16  & 44.66  & 53.71 \\
        1     & \( W_{\text{r}} = 1, W_{\text{c}} = 2 \)   & 6.73  & 9.15  & -0.25  & 0.37  & 2.81  & 6.48   & 13.77 \\
            & \( W_{\text{r}} = 2, W_{\text{c}} = 1 \)   & 7.43  & 11.43 & 12.91  & 16.11 & 3.16  & 25.63  & 40.36 \\
        \midrule
           & \( W_{\text{r}} = 0.32, W_{\text{c}} = 0.68 \)   & \textbf{13.60} & \textbf{24.43} & 30.51  & 42.22 & \textbf{4.21}  & 55.38  & 68.47 \\
        2   & \( W_{\text{r}} = 1, W_{\text{c}} = 2 \)   & 5.24  & 6.88  & 15.45  & 17.96 & 0.70  & 20.69  & 21.58 \\
            & \( W_{\text{r}} = 2, W_{\text{c}} = 1 \)   & 8.36  & 12.86 & 39.49  & 54.26 & 1.75  & 26.31  & 21.58 \\
        \midrule
         & \( W_{\text{r}} = 0.32, W_{\text{c}} = 0.68 \)   & 8.63  & 17.33 & 36.58  & 50.37 & 2.46  & 56.46  & 66.31 \\
        2.5    & \( W_{\text{r}} = 1, W_{\text{c}} = 2 \)   & 9.89  & 14.00 & 30.63  & 46.30 & 2.81  & 40.52  & 62.09 \\
            & \( W_{\text{r}} = 2, W_{\text{c}} = 1 \)   & 9.60  & 17.11 & 42.53  & 56.48 & 1.40  & 59.25  & 69.98 \\
        \midrule
           & \( W_{\text{r}} = 0.32, W_{\text{c}} = 0.68 \)   & 7.15  & 11.42 & 39.49  & 53.89 & 1.40  & 55.98  & 63.45 \\
         3   & \( W_{\text{r}} = 1, W_{\text{c}} = 2 \)   & 10.16 & 21.17 & 33.29  & 43.89 & 0.70  & 43.45  & 47.63 \\
            & \( W_{\text{r}} = 2, W_{\text{c}} = 1 \)   & 9.60  & 18.35 & 42.53  & 55.37 & 2.81  & 61.90  & 47.63 \\
        \midrule
           & \( W_{\text{r}} = 0.32, W_{\text{c}} = 0.68 \)   & 10.80 & 19.38 & \textbf{45.82}  & \textbf{58.52} & 0.70  & \textbf{65.09}  & \textbf{72.92} \\
        5    & \( W_{\text{r}} = 1, W_{\text{c}} = 2 \)   & 4.69  & 3.88  & 33.92  & 40.19 & 0.35  & 38.61  & 15.73 \\
            & \( W_{\text{r}} = 2, W_{\text{c}} = 1 \)   & 7.43  & 11.90 & 39.49  & 52.96 & 5.26  & 52.26  & 15.73 \\
        \midrule
          & \( W_{\text{r}} = 0.32, W_{\text{c}} = 0.68 \)   & 9.60  & 18.34 & 42.53  & 55.37 & 1.40  & 62.92  & 70.89 \\
        10    & \( W_{\text{r}} = 1, W_{\text{c}} = 2 \)   & 7.43  & 13.91 & 24.94  & 25.19 & 4.91  & 32.37  & 34.88 \\
            & \( W_{\text{r}} = 2, W_{\text{c}} = 1 \)   & 7.43  & 11.72 & 35.44  & 47.41 & -4.21 & 40.53  & 34.88 \\
        \bottomrule
    \end{tabular}
    }
\end{table*}
\captionsetup{font={footnotesize,rm},justification=centering,labelsep=period}%

\subsection{Ablation Study: Multi-Demographic Fairness}

The objective of our ablation study was to evaluate the model's performance when optimizing fairness across multiple demographic attributes simultaneously, specifically with respect to both \emph{country} and \emph{race}. This ablation was conducted to address \emph{RQ3}, which explores the impact of varying fairness weights for each attribute when multiple fairness attributes are considered together.

To ensure fairness, we removed these attributes from the input space, preventing the model from learning direct associations between them and the paper acceptance decisions. Instead, demographic parity loss was computed for each attribute during training, capturing deviations from fairness. The parity losses for both country and race were combined by assigning weights: \(W_c\) for country and \(W_r\) for race, with the initial weights set to \(W_c = 0.68\) and \(W_r = 0.32\), reflecting the distribution of protected groups.

To further explore the model's behavior and answer \emph{RQ3}, we varied these weights, first increasing \(W_c\) while keeping \(W_r\) constant, and then increasing \(W_r\) while keeping \(W_c\) fixed. Additionally, we experimented with different values of the fairness regularization parameter \(\lambda\), which controls the trade-off between fairness and utility. These experiments allowed us to observe how different weight configurations and fairness constraints influenced the model’s ability to achieve demographic fairness while maintaining utility and the quality of selected papers.

The results of the ablation study, shown in Table \ref{table:ablationstudy}, reveal that at \(\lambda = 1\), assigning equal weights to both race and country (\(W_r = 0.32, W_c = 0.68\)) produced significant gains for race, with a Macro Gain of 30.51\% and a Micro Gain of 46.3\%, while country showed relatively smaller improvements (6.17\% and 6.34\%, respectively). However, when the weight for country was increased (\(W_c = 2 \times 0.68\)), diversity gains for race dropped sharply, with a negative Macro Gain (-0.25\%), while country experienced slight improvements. Conversely, increasing the weight for race (\(W_r = 2 \times 0.32\)) resulted in improved diversity for both race and country, indicating that assigning more weight to race enhances diversity for both attributes to some degree.

At \(\lambda = 2.5\), the model achieved the best balance between diversity and utility. Equal weights for race and country yielded Macro and Micro Gains of 36.58\% and 50.37\% for race, and 8.63\% and 17.33\% for country, with a low utility loss of 2.46\%. This suggests that \(\lambda = 2.5\) is optimal for balancing fairness and utility. As \(\lambda\) increases further, race diversity continues to improve (reaching 45.82\% Macro Gain at \(\lambda = 5\)), but at the cost of decreasing utility. The different optimal \(\lambda\) values for race and country suggest that disparity ratios impact how fairness constraints should be weighted, with race requiring a higher \(\lambda\) due to its higher disparity ratio. This leads to greater race diversity gains at higher \(\lambda\) values, whereas country achieves optimal results at moderate \(\lambda\) values, such as 2.5.

These findings directly address \emph{RQ3}, demonstrating that fairness weights must be carefully calibrated for each protected attribute. Assigning greater weight to race tends to improve diversity for both race and country, whereas increasing the weight for country may result in reduced fairness for race. The optimal balance between fairness and utility is achieved when fairness weights and \(\lambda\) values are adjusted based on the unique disparity ratios of each attribute.

\section{Conclusion and Future Work}

This study introduces a fairness-oriented paper recommendation methodology that enhances demographic parity for race and country while maintaining academic quality. Our findings indicate that adjusting fairness requirements, including the regularization parameter \(\lambda\) and demographic weights, improves diversity while maintaining selection criteria.

Ablation experiments indicate that variations in race and country necessitate more stringent fairness requirements for optimal inclusion. Although beneficial, our technique lacks explicit causal modeling, which could enhance bias reduction. Investigating sophisticated designs such as Variational AutoEncoders (VAE) or graph-based models could enhance fairness and precision.

Incorporating institutional connections and combining causal fairness may improve bias mitigation. Confronting these obstacles will enhance fairness-oriented proposals, promoting a more inclusive peer review process.

\section*{Acknowledgment}
This work was supported by the National Science Foundation (NSF) under Award number OIA-1946391, Data Analytics that are Robust and Trusted (DART).

\bibliographystyle{IEEEtran}
\bibliography{bibfile_cleaned_cited_only}
\end{document}